\documentclass[10pt,twocolumn,letterpaper]{article}
\usepackage{cvpr}
\usepackage{times}
\usepackage{epsfig}
\usepackage{graphicx}
\usepackage{amsmath}
\usepackage{amssymb}
\usepackage{algorithm}
\usepackage{algorithmic}
\usepackage{subcaption}    
\usepackage{relsize}
\usepackage[revision]{revdiff}
\DeclareMathOperator*{\argminA}{arg\,min} % Jan Hlavacek
\DeclareMathOperator*{\minimize}{min}
\def\M{\mathcal{M}}
\def\B{\mathcal{B}}
\def\C{\mathcal{C}}
\def\N{\mathcal{N}}

\def\F{\mathcal{F}}

\def\x{\mathbf{x}}
\def\y{\mathbf{y}}

\def\bS{\mathbb{S}}
\def\bM{\mathbb{M}}
\def\bQ{\mathbb{Q}}

\newtheorem{definition}{{\bf Definition}}

\graphicspath{{figures/}}  % list here all the paths to your figure folders
\usepackage[pagebackref=true,breaklinks=true,letterpaper=true,colorlinks,bookmarks=false]{hyperref}

 \cvprfinalcopy % *** Uncomment this line for the final submission

\def\cvprPaperID{7873} % *** Enter the CVPR Paper ID here

\begin{document}

%%%%%%%%% TITLE
\title{Efficient and Robust Shape Correspondence via Sparsity-Enforced Quadratic Assignment}

\author{Rui Xiang\\ 
Department of Mathematics\\
UC Irvine\\
{\tt\small xiangr1@uci.edu}
% just like the second author
\and
Rongjie Lai\\
Department of Mathematics\\
Rensselaer Polytechnic Institute\\
{\tt\small lair@rpi.edu}
\and
Hongkai Zhao\\
Department of Mathematics\\
UC  Irvine\\
{\tt\small zhao@uci.edu}
}
\maketitle

% \thispagestyle{empty}
%%%%%%%%% ABSTRACT
\begin{abstract}
In this work, we introduce a novel local pairwise descriptor and then develop a simple, effective iterative method to solve the resulting quadratic assignment through sparsity control for shape correspondence between two approximate isometric surfaces. Our pairwise descriptor is based on the stiffness and mass matrix of finite element approximation of the Laplace-Beltrami differential operator, which is local in space, sparse to represent, and extremely easy to compute while containing global information. It allows us to deal with open surfaces, partial matching, and topological perturbations robustly. To solve the resulting quadratic assignment problem efficiently, the two key ideas of our iterative algorithm are: 1) select pairs with good (approximate) correspondence as anchor points, 2) solve a regularized quadratic assignment problem only in the neighborhood of selected anchor points through sparsity control. These two ingredients can improve and increase the number of anchor points quickly while reducing the computation cost in each quadratic assignment iteration significantly. With enough high-quality anchor points, one may use various pointwise global features with reference to these anchor points to further improve the dense shape correspondence. We use various experiments to show the efficiency, quality, and versatility of our method on large data sets, patches, and point clouds (without global meshes).

\end{abstract}

%%%%%%%%% BODY TEXT
\section{Introduction}
Geometric modeling and shape analysis is ubiquitous in computer vision, computer graphics, medical imaging, virtual reality, 3D prototyping and printing, data analysis, etc. Shape correspondence is a basic task in shape registration, comparison, recognition, and retrieval. Unlike images, shapes do not have a canonical representation domain or basis and do not form a linear space. Moreover, their embedding can be highly ambiguous even for intrinsically identical ones. Further complications in practice include noise, topological perturbations (holes), partial shapes, and lack of a good triangulation. These difficulties pose both modeling and computational challenges for shape modeling and analysis. 

For dense shape correspondence, the first step is to design desirable descriptors, pointwise, or pairwise. Pointwise descriptors can be extrinsic and local (in space) \cite{tombari2010unique, gumhold2001feature, dubrovina2011approximately, rodola2012game}, or intrinsic (invariant under isometric transformation). Extrinsic pointwise descriptors usually have difficulties in producing accurate dense correspondence, especially if there is non-rigid transformation involved. Many intrinsic pointwise descriptors in the space domain, such as geodesics distance signatures \cite{van2011survey}, heat kernel signatures \cite{sun2009concise}, wave kernel signatures \cite{aubry2011wave}, and in spectrum domain using eigen-functions of the Laplace-Beltrami operator (LBO) have been proposed~\cite{Reuter:06,Levy:2006IEEECSMA,Vallet:2008CGF,Bronstein:2010CVPR,lai2010metric,lai2017multiscale}. For example, functional maps \cite{ovsjanikov2012functional} aims to find proper linear combinations of truncated basis functions, e.g., eigen-functions of the Laplace-Beltrami operator, based on some prior knowledge, e.g., given landmarks and/or region correspondence, as the pointwise descriptor. Then various nearest neighbor searching or linear assignment methods are used in the descriptor space to find the dense point correspondence. These intrinsic pointwise descriptors are typically nonlocal and require to solve certain partial differential equations, e.g., the Laplace-Beltrami equation, on a well-triangulated mesh. Hence they can be sensitive to topological perturbations and boundary conditions. Moreover, pointwise descriptors based on a truncated basis in the spectrum domain lose fine details in the geometry.
 On the other hand, using good pairwise descriptors, such as pairwise geodesic distance matrix\cite{vestner2017product} or kernel functions \cite{vestner2017efficient}, to find shape correspondence is usually more robust and accurate since the matching needs to satisfy more and stricter constraints to minimize some kind of distortion. However, a very challenging computational problem, a quadratic assignment problem (QAP) which is NP-hard, needs to be solved \cite{lawler1963quadratic}. Different kind of methods have been proposed to solve the QAP approximately in a more computational tractable way e.g sub-sampling \cite{tevs2011intrinsic}, coarse-to-fine \cite{wang2011discrete}, geodesic distance sparsity enforcement methods \cite{gasparetto2017spatial} and various relaxation approaches \cite{aflalo2015convex, bronstein2006generalized, kezurer2015tight, leordeanu2005spectral, chen2015robust,dym2017ds++}. 
One popular approach is to relax the nonconvex permutation matrix (representing pointwise correspondence) constraint in the QAP to a doubly stochastic matrix (convex) constraint \cite{aflalo2015convex,chen2015robust}. However, both the pairwise descriptor and the doubly stochastic matrix are dense matrices, which causes the relaxed QAP still challenging to solve even for a modest size problem. 

\begin{figure}
\centering
\includegraphics[width=\linewidth]{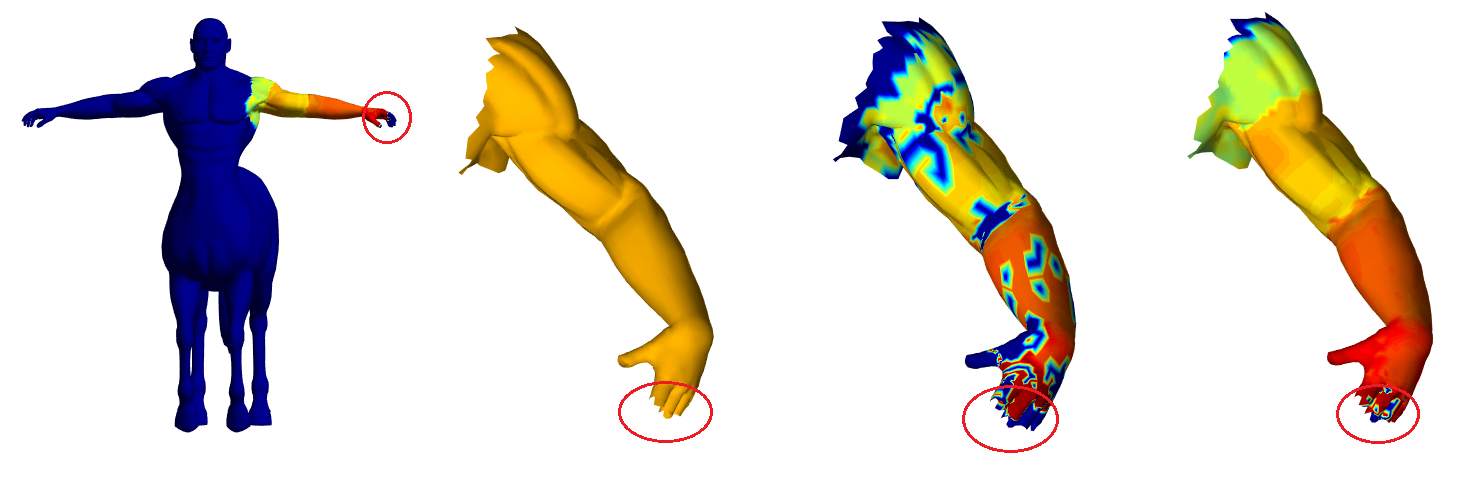}
\caption{Example of partial matching with topological changes. Topological changes are highlighted by red circles. The patch in second column is mapped onto the entire shape in the first column, and non-blue area is the ground truth map. Extra points in the entire shape are colored in blue. The third column is the mapping result using SHOT, and the last column is the mapping result from our method.}
\label{patch3}
\end{figure}

In this work, we propose a novel approach for dense shape correspondence for two nearly isometric surfaces based on local pairwise descriptor and an efficient iterative algorithm with sparsity control for the doubly stochastic matrix to solve the relaxed QAP. The main novelty and contribution of our proposed method include:

\noindent 1) A local pairwise descriptor using the combination of the stiffness (corresponding to the finite element approximation of the LBO) and the mass matrix (corresponding to local area scaling). It only involves interactions among local neighbors and is extremely simple to compute. Note that all local interactions are coupled like heat diffusion through the whole shape. In other words, global and full spectral information of LBO is embedded implicitly in our pair-wise descriptor. Due to the locality, the descriptor enjoys stability and good performance for open surfaces and with respect to topological perturbations, as shown in Figure \ref{patch3} and by more examples in Section \ref{sec:result}. The sparsity of the pairwise descriptor also reduces the computation cost for the relaxed QAP.

\noindent 2) An efficient iterative algorithm with sparsity control for the resulting relaxed QAP. We first use a local distortion measurement (see Section \ref{sec:faith} for details) to select pairs from both shapes with good correspondence as anchor points for the next iteration. Using regularity of the map, we enforce that the neighborhood of anchor points can only map to the neighborhood of the corresponding anchor points which induces a sparsity structure in the doubly stochastic matrix. It results in a significant reduction of variables and hence, the computation cost in each iteration. As we demonstrate in the numerical experiments, the number of high-quality anchor points grow quickly with iterations.

Here is the outline of our paper. We introduce our quadratic assignment model based on a local pairwise descriptor in Section~\ref{sec:QAP} and then present an efficient iterative algorithm to solve the quadratic assignment problem with sparsity control in Section~\ref{sec:iterative}. In Section~\ref{sec:dis}, we extend our method to point cloud data and patch matching. Numerical experiments are demonstrated in Section~\ref{sec:result} and conclusion follows. 

% \textcolor{red}{RX: add a sentence for section 4?}
%------------------------------------------------------------------------
\section{Quadratic Assignment Model Using Local Pairwise Descriptors} 
\label{sec:QAP}
Given two manifolds $\M_1$ and $\M_2$ sampled by two point clouds $\mathcal{P}_1=\{\x_i\}_{i=1}^{n}$ and $\mathcal{P}_2=\{\y_i\}_{i=1}^{n}$ respectively, the typical task of dense shape correspondence is to find a point-to-point map between $\mathcal{P}_1$ and $\mathcal{P}_2$. Let $\bQ_1\in \mathbb{R}^{n\times n}$ and $\bQ_2\in \mathbb{R}^{n\times n}$ be two given pairwise descriptors, e.g., pairwise geodesic distance, between two points in $\mathcal{P}_1$ and $\mathcal{P}_2$ respectively. The shape correspondence problem can be casted as the following QAP: 
\begin{equation}
\argminA_{P \in \Pi_{n}} \|P \bQ_1 - \bQ_2 P\|_F^2
\label{eq:QAP}
\end{equation}
where $P\in \mathbb{R}^{n\times n}$ is a permutation matrix with binary $\{0,1\}$ elements and each row and column sum is 1, and $\|\cdot\|_F$ is the Frobenius norm of a matrix. 

Since the QAP problem is NP-hard~\cite{sahni1976p}, it is common to relax the permutation matrix in \eqref{eq:QAP} to a doubly stochastic matrix, $D\in \mathcal{D}_n = \{D \in \mathbb{R}^{n \times n}|\ D \vec 1=\vec 1,D^{\mathsmaller T} \vec 1=\vec 1, D_{ij}\ge 0\}$, in the shape registration context~\cite{aflalo2015convex, bronstein2006generalized, kezurer2015tight, leordeanu2005spectral}. The doubly stochastic matrix representation not only convexifies the original QAP~\eqref{eq:QAP} but also provides a more general probabilistic interpretation of the map. However, there remain at least two major computational challenges to solve the relaxed QAP for correspondence problems between shapes of relatively large size. First, the usual choice of pairwise descriptors, such as pairwise distance \cite{vestner2017product}, heat kernel~\cite{bronstein2010scale}, and wave kernel~\cite{aubry2011wave} are represented as dense matrices and so are the doubly stochastic matrix. It can pose a storage and memory issue when two shapes are of large size even before conducting any computation. In this case, certain approximation has to be used, such as sub-sampling methods \cite{tevs2011intrinsic}, truncation of pairwise descriptors or spectrum approximation \cite{aflalo2016spectral}, though they may lead to accuracy problems due to the approximation error. 
Computationally, single dense matrix multiplication of the pairwise descriptor matrix and the doubly stochastic matrix requires $O(n^3)$ operations, where $n$ is the number of points. More seriously, the relaxed QAP is usually solved by an iterative method. Due to the coupling of all elements of the doubly stochastic matrix, i.e., every element is affected by all other elements, elements corresponding to good matching can be influenced by those corresponding to the wrong matching initially, which can cause a slow convergence of the optimization process  especially when the initial guess is not good enough. Furthermore, for data with noise or distortion, the QAP may propagate the distortion or noise in one region to other regions and cause the solution to the QAP unsatisfactory.

%updating the whole doubly stochastic matrices at each iteration will most likely propagate the error in the distorted region to other regions and cause the solution of the optimization model unsatisfactory results.   %Lastly, the permutation matrix recovered from the doubly stochastic matrix may be far away from the true solution due to the relaxation alongside noise in real data.   

To tackle the aforementioned challenges for the QAP, we propose the following relaxed quadratic assignment using sparse pairwise descriptors and develop an efficient iterative algorithm with sparsity control for the doubly stochastic matrix to find high-quality dense landmarks. These landmarks are then used in the final post-processing step to construct the full correspondence.    

%Moreover, we explore a new algorithm to solve the proposed problem by dynamically enforcing sparsity of the doubly stochastic matrices using a region growing based method. This can be viewed as we geometrically regularize the sparsity of doubly stochastic matrices such that the resulting correspondence satisfying a 'smoothness' property, i.e., nearby points on the source surface should be mapped to nearby points on the target surfaces. 
%------------------------------------------------------------------------

\subsection{Sparse pairwise descriptors}
Let $(\M,g)$ be a closed 2-dimensional Riemannian manifold, the LBO  is defined as $\displaystyle \Delta_{(\M,g)} \psi = \frac{1}{\sqrt{G}} \sum_{i=1}^2 \frac{\partial}{\partial x_i}(\sqrt{G} \sum_{j=1}^2 g^{ij} \frac{\partial \psi}{\partial x_j})$ \cite{chavel1984eigenvalues}, here $g^{ij}$ is the inverse of $g_{ij}$ and $G = det(g)$. 
LBO is an elliptic and self-adjoint operator intrinsically defined on the manifold; thus, it is invariant under isometric transformation. The LBO eigen-system satisfies:
\begin{equation}
    \Delta_{(\M,g)}\psi_i = -\lambda_i \psi_i,\quad \int_{\M}\psi_i \psi_j \mathrm{d} s = \delta_{ij}
\label{lbo}
\end{equation}
and  uniquely determines the underlying manifold up to isometry~\cite{berard1994embedding}. Spectral geometry is widely used in shape analysis ~\cite{Reuter:06,Levy:2006IEEECSMA,Vallet:2008CGF,sun2009concise, Bronstein:2010CVPR,lai2010metric,ovsjanikov2012functional,lai2017multiscale,schonsheck2018nonisometric}. 

In practice, $\M$ is discretized by a triangular mesh $T = \{\tau_\ell\}$ with vertices $V = \{\x_i\}_{i=1}^n$ connected by edges $E=\{e_{ij}\}$. For each edge $e_{ij}$ connecting points $p_i$ and $p_j$, we define the angles opposite $E_{ij}$ as angles $\alpha_{ij}$ and $\beta_{ij}$. Denote the stiffness matrix as $\bS$, given by \cite{reddy1993introduction,reuter2009discrete}
\begin{equation}
\bS_{ij}= 
\begin{cases}
-\frac{1}{2} [cot \alpha_{ij} + cot \beta_{ij}]& \quad i\sim j\\
\sum_{k \sim i} \bS(i,k)& \quad  i = j
\end{cases}
\end{equation}
where $\sim$ denotes the connectivity relation by an edge. The mass matrix $\bM$ is given by 
\begin{equation}
\bM_{ij}=
\begin{cases}
\frac{|\tau_1|+|\tau_2|}{12} &i\sim j\\
\sum_{k\sim i} \bM(i,k)& {i=j}
\end{cases}
\end{equation}
where $|\tau_1|$ and $|\tau_2|$ are the areas of the two triangles sharing the same edge $ij$. On the one hand, the eigensystem of LBO can be computed as $\bS \psi = \lambda \bM \psi $, which suggests $\bS$ and $\bM$ implicitly contain the spectrum information of LBO which can be used to determine a manifold uniquely up to isometry. On the other hand, 
it has been rigorously shown a global rigidity theorem on the Stiffness matrix, i.e. two polyhedral surfaces share the same Stiffness matrices if on only if their corresponding metrics are the same up to a scaling~\cite{gu2010discrete}. Note that the  mass matrix fixes the scaling factor. Furthermore, both of these two matrices are local which are not sensitive to boundary conditions or topological perturbations. Therefore, we expect that $\bS$ and $\bM$ together can serve as good sparse pairwise descriptors in a QAP formulation for shape correspondence. 

%Then \eqref{lbo} can be discretized into following linear system:
%\begin{equation}
%    Sf = -\lambda M f,\ f:=(f(p_i))_{i=1}^n
%    \label{dislbo}
%\end{equation}

%Computing eigen-functions and eigen-values of LBO are fast. However, truncation error and sign ambiguity of eigen-functions will always be introduced during the computation. 

%Also LBO is very sensitive to boundary conditions and topological changes of the mesh which limits its usage on shape with boundaries or with topological changes. Based on the fact that, LBO actually comes from \eqref{dislbo}, stiffness matrix and mass matrix already capture the intrinsic properties of these two manifolds; and furthermore, these two matrices are local which are not sensitive to boundary condition or small local topological changes. So we formulated our idea of directly align stiffness matrix and mass matrix.
%------------------------------------------------------------------------

\subsection{Relaxed QAP for shape correspondence}
Given two surfaces $\M_1$ and $\M_2$ discretized by triangular meshes with vertices $\{\x_i\}_{i=1}^n$ and $\{\y_i\}_{i=1}^n$ respectively. We denote the corresponding stiffness matrices by $\bS_1, \bS_2$ and the corresponding mass matrices by $\bM_1$ and $\bM_2$. Representing a point-to-point mapping between $\M_1$ and $\M_2$ by a permutation matrix $P\in \Pi_{n}$, we propose the following QAP problem to construct the point-to-point mapping between these two surfaces: 
\begin{equation}
    \minimize_{P \in \Pi_{n}}\ \frac{1}{2} \|P \bS_1-\bS_2 P\|_F^2 +  \frac{\mu}{2} \|P \bM_1 - \bM_2 P\|_F^2,
\label{eqn_QAPSM} 
\end{equation}
Where $\mu$ is a balance parameter. The stiffness matrix captures local geometric information, and the mass matrix captures local area information of the discretized surface. Both matrices have a sparsity structure with the number of nonzero entries linearly scaled with respect to the number of points. This nice sparse property of both matrices already alleviates the memory issue for large data sets significantly. In addition, since both descriptors only capture local geometric information, it potentially allows the proposed model to handle partial matching problem, open surfaces, and topological changes. 
%The global information of each point can also be obtained in a propagation sense as all points are directly or indirectly connected. Most important, they are both sparse. 

Since the proposed QPA is NP-hard, we relax the permutation matrix to a doubly stochastic matrix representation of the mapping:
%$D\in \mathcal{D}_n = \{D \in \mathbb{R}^{n \times n}|\ D \vec 1=\vec 1,D^{\mathsmaller T} \vec 1=\vec 1, D_{ij}\ge 0\}$. 
%This leads to an optimization problem as follows:
\begin{equation}
    \minimize_{D \in \mathcal{D}_n}\ \frac{1}{2}\|D \bS_1-\bS_2 D\|_F^2  + \frac{\mu}{2} \| D \bM_1 - \bM_2 D \|_F^2 
\label{eqn_QAPSM_relaxed}
\end{equation}
As an advantage of this relaxation, each row of $D$ can be interpreted as the probability of a point on $\M_1$ mapping to points on $\M_2$. Now the relaxed QAP \eqref{eqn_QAPSM_relaxed} is convex and can be solved by well-known algorithms in convex programming. Here, we use projected gradient descent algorithm with Barzilai-Borwein step size solve this optimization problem (see details in Section \ref{subsec:alg}).

\section{Dynamically sparsity-enforced QAP}
\label{sec:iterative}
As we pointed out before, the relaxed QAP problem \eqref{eqn_QAPSM_relaxed} is still difficult to solve if dense doubly stochastic matrices are used in the optimization process. To overcome those difficulties, we propose an iterative algorithm that 1) selects candidates for well-matched pairs as anchor points, 2) enforces a dynamic sparsity structure of the doubly stochastic matrix by using regularity of the map, i.e., nearby points on the source surface should be mapped to nearby points on the target surface, in the neighborhood of those paired anchor points in each iteration. These two ingredients both reduce the computation cost in each iteration (only sparse matrices are involved) and increase the number of well-matched pairs quickly since only candidates for well-matched points are used to guide the iterations.

%limitations, we impose the following idea of dynamically regularizing the sparsity structure in the optimization procedure by assuming certain `smoothness' of the desired mapping, namely, we prefer nearby points on the source surface should be mapping to nearby points on the target surface. Besides, if we can fix the map at points which are mapped correctly during optimization, it will not only help reduce the computation cost but also increases the accuracy. 

\subsection{Local distortion test}
\label{sec:faith}
To define a desired sparsity structure for the doubly stochastic matrix $D$ in the relaxed QAP~\eqref{eqn_QAPSM_relaxed}, we first need to detect candidates for well-matched pairs, or equivalently to remove those definitely ill-matched points, dynamically in each iteration. Motivated by the Gromov-Wasserstein distance \cite{memoli2011gromov} and the unsupervised learning loss in \cite{halimi2018self}, we introduce the following criterion to quantify location distortion of a mapping at a point on the source manifold. 
\begin{definition}[Local distortion criterion]
\label{def:faithfulness}
Let $\phi:\M_1 \rightarrow \M_2$ be a map between two isometric manifolds. For any point $\x\in\M_1$, consider its $\gamma$-geodesic ball in $\M_1$ as $\B_\gamma(\x) = \{\y\in\M_1~|~ d_{\M_1}(\x,\y)\leq \gamma\}$. local distortion of $\phi$ at $\x$ is defined as:
\begin{equation}
    \F_{\gamma}(\phi)(\x) = \frac{1}{|\B_\gamma(\x)|} \int_{\y\in \B_\gamma(\x)} DE_\phi(\x,\y)\mathrm{d} \y 
\label{eq:faithfulness}
\end{equation}
where 
%$\displaystyle DE_\phi(\x,\y) = \frac{|d_{\M_1}(\x,\y)-d_{\M_2}(\phi(\x),\phi(\y))|}{d_{\M_1}(\x,\y)} $ 
$\displaystyle DE_\phi(\x,\y) = \frac{1}{\gamma}|d_{\M_1}(\x,\y)-d_{\M_2}(\phi(\x),\phi(\y))|$ 
is the difference between the geodesic distance $d_{\M_1}, d_{\M_2}$ on the two corresponding manifolds, and $|\B_\gamma|$ is the volume of $\B_\gamma$. 
% We intentionally use $|\B_\gamma|^{3/2}$ to normalize the unit of the above integral. 
% local distortion of $\phi$ at $\x$ is defined as:
% \begin{equation}
%     \F_{\gamma}(\phi) = \frac{1}{|\M_1|} \int_{\x\in \M_1} \F(\phi)(\x) \mathrm{d} \x
% \end{equation}
\end{definition}

We have the following straightforward properties:
\begin{enumerate}
    \item 
    If $\phi$ is an isometric map, $\F_{\gamma}(\phi)(\x) = 0, \forall \x\in\M_1, \gamma>0$.
    \item
    If $\F_{\gamma}(\phi)(\x) = 0, \forall \x\in\M_1$ for some $\gamma>0$, $\phi$ is isometric.
\end{enumerate}
In discrete setting, $\M_1$ is represented as $ \{\x_i\}_{i=1}^n$, $\M_2$ is represented as $ \{\y_i\}_{i=1}^n$and the map $\phi$ is discretized as a one-to-one map between $ \{\x_i\}_{i=1}^n$ and $ \{\y_i\}_{i=1}^n$. We use the following discrete approximation: %\RJ{I think we should use right approximation to handle shapes with different points} 
\begin{equation}
     \F_{\gamma}(\phi)(\x_i) \approx \frac{\sum_{\x_j \in \B_\gamma(\x_i), \x_j\ne \x_i} \bM_1(j,j) DE_\phi(\x_i,\x_j)}{  \left(\sum_{\x_j \in \B_\gamma(\x_i), \x_j\ne \x_i } \bM_1(j,j) \right)}
    \label{eq:test} 
\end{equation}
to quantify how much $\phi$ is distorted locally and use it to prune out those points that have large local distortion in the next iteration for the QAP. In practice, we specify a number $\epsilon$ and view $x_i$ as a candidate of well-matched anchor point for $\phi$ if $\F(\phi)(\x_i) < \epsilon$. Together with $\phi(\x_i)$, we extract a collection of anchor pairs $\{(\x_i,\phi(\x_i))\}_{i=1}^k$ which are used to define sparsity pattern in the doubly stochastic matrix $D$ dynamically in the relaxed QAP \eqref{eqn_QAPSM_relaxed}. It is important to note that current anchor pairs will be re-evaluated and updated in later iterations.

\subsection{Dynamic sparsity for doubly stochastic matrices}
Suppose a collection of anchor pairs $\{(\x_i,\phi(\x_i))\}_{i=1}^k$ have been selected using the local distortion test. In the next iteration, a sub-QAP only involving points in the neighborhood of selected anchor pairs are solved. We further enforce a sparsity structure on the doubly stochastic matrix for the sub-QAP based on the following two rules.
\begin{enumerate}
    \item Each anchor point is mapped to its corresponding anchor point;
    \item Points in the neighborhood of an anchor point on the source surface are mapped to the neighborhood of the corresponding anchor point on the target surface. 
\end{enumerate}
Let $\N(\x)$ denote the neighborhood of a given point on a manifold, e.g., a geodesic ball $\B_r$ centered at $\x$ on the manifold, or simply points in the $l$-th ring of $\x$ on a triangular mesh. 
%We remark that we always choose the neighborhood size larger than that for distortion test. 
Define $\N(\{\x_i\}_{i=1}^k) = \bigcup_{i=1}^k \N(\x_i)$ and $\N(\{\phi(\x_i)\}_{i=1}^k) = \bigcup_{i=1}^k \N(\phi(\x_i))$. For the doubly stochastic matrix $D$ in the relaxed QAP~\eqref{eqn_QAPSM_relaxed}, we only update variables with indices corresponding to the set $\N(\{\x_i\}_{i=1}^k)\times \N(\{\phi(\x_i)\}_{i=1}^k)$ together with the following sparsity constraints 
\begin{equation}
    D_{t,s} = 
    \begin{cases}
    \delta_{\phi(\x_s),\y_t}, & \text{if}~\x_s\in \{\x_i\}_{i=1}^k \\
    0,&  \text{if}~\x_s\in \N(\x_i) ~ \text{and}~ \y_t\notin \N(\phi(\x_i)) \\
    0, & \text{if}~\y_t\in \N(\phi(\x_i)) ~ \text{and}~ \x_s\notin \N(\x_i)
    \end{cases}.
\label{eqn_sparsity}
\end{equation}

By limiting the optimization region and enforcing the previous two sparsity constraints, the number of variables in the QAP problem after the sparsity enforcement is greatly reduced from $O(n^2)$ to $O(n)$. This can dramatically reduce computation cost. Moreover, since the anchor points are fixed, it will no longer be influenced by other points in the current optimization process; on the contrary, it will enforce a positive influence on the neighboring points.

In practice, we always choose the size of $\B_\gamma(\x)$ in the distortion test smaller than the size of sparsity control neighborhood $\N(\x)$ to allow the growth of anchor points in the next iteration. In our experiments, we choose $\B_\gamma(\x_i)$ as points included in the second ring of $\x_i$ and $\N(\x_i)$ as points included in the fourth ring of $\x_i$. The larger $\B_\gamma$ is, the more precise anchor points will be; the larger sparsity neighborhood $\N(\x)$ is, the faster the number of anchor points grows. However, computation cost also increases for each QAP iteration when $\B_\gamma(\x)$ and $\N(\x)$ become larger since the doubly stochastic matrix is less sparse. 

\begin{figure}[htbp]
\centering
\includegraphics[width=\linewidth]{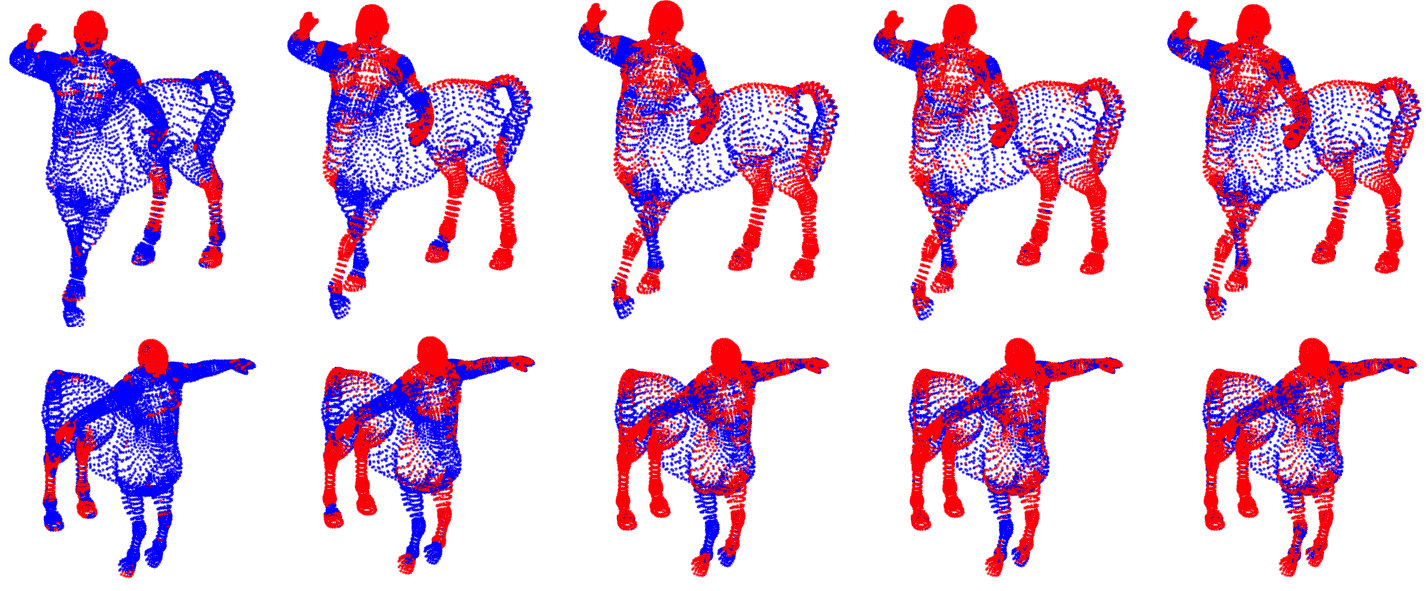}
\caption{Growth of anchor pairs. Red points in each column from left to right represent the location of anchor pairs from iteration 1 to 5. %The update on the sparsity regularization serves as a bridge from double stochastic matrix to permutation matrix. It makes the optimization solvable meanwhile the solution not too far away from the ground truth permutation
}
\label{fig_anchorpairs}
\end{figure}

Once the sparsity regularized $D$ is obtained, we update the point-to-point mapping $\phi$ by choosing the largest element in each row. Then, we find a new collection of anchor pairs by the distortion test based on the updated $\phi$. Figure \ref{fig_anchorpairs} illustrates an example of this procedure in the first 5 iterations.  
Ideally, one should grow anchor points until all points are covered. However, because of noise and/or non-isometry, 
the growth of high-quality anchor points usually slows down after a few iterations. 
%two point clouds describing two isometric manifold may not have exactly the same stiffness matrix and mass matrix up to a permutation.
Moreover, even the exact solution of QAP \eqref{eq:QAP} may not produce a desirable result. 
%The growth of high anchor points in our method will slow down as the iteration number increases. %For some shape with relatively large noise, growth of anchor points may stop before covering all points. 
To balance between efficiency and accuracy, we find that 5 iterations of relaxed QAP~\eqref{eqn_QAPSM_relaxed} is good enough to find enough high quality anchor points. We then use a post-processing step to construct the correspondence for the remaining points with the help of matched anchor pairs. 

%\RJ{Reviewers may question us why ignore QAP but only use linear programming eventually. One possible way we can add is to let anchor points cover the whole manifold and check the faithfulness error, if it less than certain number then use linear programming. Even though, we also need to carefully explain why eventually use linear programming. For instance, they may ask why do not we use similar sparsity enforcement in linear programming.}

\subsection{Post Processing}
\label{subsec_postprocessing}
The aforementioned sparsity enforced quadratic assignment model based on local features is effective in growing anchor points from initial seed points in the regions where there is no significant non-isometric distortion. However,  for regions where there is significant distortion, such as near fingertips or elbow regions of humans in different poses, local features may not be enough to produce satisfactory results. To construct a full correspondence and improve mappings in those regions, one can use global pointwise descriptors with reference to those already well-matched anchor pairs. There are various options for the final post processing step. For example, we use Heat Kernel Signature (HKS)~\cite{sun2009concise} for closed surfaces in our experiments.
Let $H_1(\x,\x',t)$ and $H_2(\y,\y',t)$ denote HKS on $\M_1$ and $\M_2$ respectively. Given $\{\x_i,\phi(\x_i)\}_{i=1}^\ell$ as the (sub)set of high quality anchor pairs obtained from solving the QAP~\eqref{eqn_QAPSM_relaxed}, we construct pointwise descriptors of length $\ell$ for those $\x\in \M_1, \y\in \M_2$ not in the anchor pair set as $\{H_1(x,\x_i,t)\}_{i=1}^\ell, \{H_2(y,\phi(\x_i),t)\}_{i=1}^\ell$.  Then we simply perform a nearest neighborhood search in this descriptor space to find the correspondence for non-anchor points. For patches, we use geodesic distance to the chosen anchor pairs on the corresponding surfaces as the pointwise descriptor.

\subsection{Numerical Algorithms}
\label{subsec:alg}
 We use projected gradient descent with Barzilai-Borwein step size \cite{barzilai1988two}, summarized in Algorithm~\ref{alg1}, with the dynamic sparsity constraints~\eqref{eqn_sparsity} in each iteration to solve \eqref{eqn_QAPSM_relaxed}. 
% This algorithm always converge since \eqref{eqn_QAPSM_relaxed} is convex \RJ{ref? A better way is to try FISTA; sine it's quadratic?} . 
The initial doubly stochastic matrix $D_0$ can be a random matrix or using the initial guess provided by SHOT feature \cite{tombari2010unique} satisfying the sparsity constraint by projection~\eqref{eq:projection}. SHOT feature only need to be computed once at the very beginning to provide the initial doubly stochastic matrix $D_0$ and select anchor points for the first iteration. In later iterations, initial guess can be provided by projecting $D$ from previous iteration according to the new sparsity constraint.

\begin{algorithm}[h]                       % HERE!!!!!!!!!
\caption{Projected gradient decent for \eqref{eqn_QAPSM_relaxed}}           % give the algorithm a caption
\begin{algorithmic}
\small  % enter the algorithmic environment
%\STATE {Solve equation \eqref{eqn_QAPSM_relaxed} with sparsity constraint \eqref{eqn_sparsity} with a given initial guess $P_0$}
\REPEAT
\STATE 1.$Y_{k+1}=D_k - \alpha_k \nabla_{D} (\| D_k \bS_1-\bS_2 D_k\|_F^2+\mu \| D_k \bM_1 - \bM_2 D_k\|_F^2)$
\STATE 2.$D_{k+1}=\argminA_{D \in \mathcal{D}_n} \| D - Y_{k+1}\|^2_F$
\UNTIL{} \\
\end{algorithmic}
\label{alg1}
\end{algorithm}
Note that we only update entries of $D$ corresponding to those points in the neighborhood of selected anchor pairs $\N(\{\x_i\}_{i=1}^k)\times \N(\{\phi(\x_i)\}_{i=1}^k)$ and perform the projection on the set of doubly stochastic matrix $D$ satisfying the sparsity constraint \eqref{eqn_sparsity}. Let $\C$ be the indicator matrix for the sparsity constraint 
\begin{equation}
    \C_{t,s} = 
    \begin{cases}
    \delta_{\phi(\x_s),\y_t}, & \text{if}~\x_s\in \{\x_i\}_{i=1}^k \\
        1,&  \text{if}~\x_s\in \N(\x_i) ~ \text{and}~ \y_t\in \N(\phi(\x_i)) \\
    % 0, & \text{if}~\y_t\in \N(\phi(\x_i)) ~ \text{and}~ \x_s\notin \N(\x_i) \\
    0, & \text{otherwise}
    \end{cases}.
\end{equation}
The solution to the projection step in Algorithm \ref{alg1} 
\begin{equation}
    D_{k+1} = \argminA_{D \in \mathcal{D}_n} \| D - Y\|^2_F,\ \text{s.t. } \eqref{eqn_sparsity}
\end{equation}
is given by
\begin{equation}
    \begin{aligned}
        D_{k+1} &= \Big(Y + \frac{|Y_{\C}| - |\C|}{|\C|^2} \vec{1}  \vec{1}^{\mathsmaller T}\\
        &- (Y_{\C}^{\mathsmaller T} \vec{1} - \vec{1}) \oslash \vec{c}  \vec{1}^{\mathsmaller T} - \vec{1}  ((Y_{\C}  \vec{1} - \vec{1} ) \oslash \vec{r})^{\mathsmaller T}\Big)_{\C}
    \end{aligned}
\label{eq:projection}
\end{equation}
where $(\boldsymbol{\cdot})_{\C} = (\boldsymbol{\cdot} ) \odot C$, $|\boldsymbol{\cdot} | = \vec{1}^{\mathsmaller T}  (\boldsymbol{\cdot} ) \vec{1}$, $\C \vec{1} = \vec{r}$, $\C^{\mathsmaller T} \vec{1} = \vec{c}$ and $\odot$, $\oslash$ are Hadamard product and Hadamard division. We further relax our problem by neglecting the nonegative constraint as suggested in \cite{aflalo2014graph}. This strategy further reduces the computation cost without causing any problem in all of our experiments. 

Our iterative method for the relaxed QAP~\eqref{eqn_QAPSM_relaxed} is summarized in Algorithm~\ref{alg2}. Starting from an initial point-to-point map $\phi^0$ (or converted from an initial doubly stochastic matrix), the three steps in each iteration are: \textcircled{1} Update the set of anchor pairs using ~\eqref{eq:test}; \textcircled{2} Update the doubly stochastic matrix by Algorithm~\ref{alg1} with sparsity constraint based on updated anchor pairs; \textcircled{3}     Convert the doubly stochastic matrix to an updated point-to-point map by choosing the index of the largest element in each row.

%\begin{enumerate}
%    \item 
%    Update the set of anchor pairs using ~\eqref{eq:test}.
%    \item
%    Update the doubly stochastic matrix by Algorithm~\ref{alg1} with sparsity constraint based on updated anchor pairs.
%    \item
%    Convert the doubly stochastic matrix to an updated point-to-point map by choosing the index of the largest element in each row.
%\end{enumerate}
Note that all anchor pairs are updated and improved (by decreasing local distortion tolerance $\epsilon$) during the iterations. Geometrically, our iterative method is like matching by region growing from anchor pairs. The local distortion criterion allows us to efficiently  and  robustly  select  a  few  reasonably  good  anchor points from a fast process (but not necessarily accurate dense correspondence), such as SHOT. Then anchor pairs will grow as well as improve due to gradually diminishing local distortion tolerance during iterations. In our experiments, we find enough high-quality anchor pairs after 5 iterations by decreasing $\epsilon$ from 5 to 1. Then we use these anchor pairs to construct the correspondence of remaining points in the final post-processing step as described in Section~\ref{subsec_postprocessing}.

\begin{algorithm}[h]
\caption{Iterative method for relaxed QAP with dynamic sparsity control}         
\begin{algorithmic}
\small 
\STATE \textbf{Input} a point-to-point map $\phi^0$, iteration steps n, $\{\epsilon_i\}_1^n$ and parameter $\mu$.\par
\REPEAT
\STATE 1. Find anchor pairs $\{(\x_i,\phi^k(\x_i)) ~|~ \F(\phi^k)(\x_i)< \epsilon_k  \}$. Define $\N^k_1 = \N(\{\x_i\}_{i=1}^k)$ and  $\N^k_2 = \N(\{\phi^k(\x_i)\}_{i=1}^k)$.
\STATE 2. Compute $D^{k+1}$ by Algorithm \ref{alg1} with sparsity constraint \eqref{eqn_sparsity} on $\N^k_1\times \N^k_2$.
\STATE 3. Update $\phi^{k+1}(x_s) = y_t$, where $t =\arg\max D^{k+1}(s,:)$. %is the index of the largest element in the $s$-th row of $D^{k+1}$  

\UNTIL{n steps are reached} \\
%\STATE If there are points uncovered, perform the post processing described in section \ref{subsec_postprocessing}.
% \RJ{how to chose $\phi^0$ and $\epsilonon$? added in alg 2 and previous paragraph}
\end{algorithmic}
\label{alg2} 
\end{algorithm}

Since  we  start  with  a  relatively  large local  distortion  tolerance  for  initial  anchor  pairs,  our method is quite stable with respect to the initialization. Moreover, as we decrease the tolerance with iterations, anchor pairs selected earlier can be updated in later iterations. We remark that the above algorithm can be straightforwardly extended to shape correspondence between two point clouds with different sizes by using a rectangular doubly stochastic matrix with the right dimension.

\section{Discussion}
\label{sec:dis}
%\paragraph{Projection $D$ on Birkhoff polytope with sparsity enforcement}
%\RJ{write down the projection solution. Need to say row sum equal to one on the sparsity controlled region. No need to write down details of Larangian, KKT et al. They are standard.} 

%\textcolor{red}{RX: I put it after algorithm 2}

\noindent\textbf{Point cloud matching}~~
We can easily extend our method to point clouds, raw data in many applications, without a global triangulation by constructing the stiffness and mass matrices at each point using the local mesh method \cite{lai2013local} with an adaptive-KNN algorithm. 

In \cite{lai2013local}, the local connectivity of a point $p$ on the manifold $M$ is established by constructing a standard Delaunay triangulation in the tangent plane at $p$ of the projections of its K nearest neighbors. However, the classical KNN with fixed $K$ is not adaptive to local geometric feature size or sampling resolution, which may lead to a loss of accuracy. So we introduce the following adaptive-KNN.

%For a neighborhood (patch) $\N(i)$ of $p_i$ formed by traditional KNN, the local normalized co-variance matrix of this patch is defined as 
%$\displaystyle \frac{1}{|\N(i)|} \sum_{p_j \in \N(i)} (p_j - c)^{\mathsmaller T}(p_i-c)$.
%Here $\displaystyle c = \frac{1}{|\N(i)|} \sum_{p_j \in \N(i)} p_j$ is the local centroid. 
%Let $(e_1^i,e_2^i,e_3^i)$ be the eigenvectors, which form a local orthogonal frame for the patch, and 
Let $\lambda_1^i \geq \lambda_2^i \geq \lambda_3^i$ be the corresponding eigenvalues the local normalized co-variance matrix. The key idea of our adaptive-KNN is that the local patch should not deviate too much from a planar one for a good linear approximation of the local geometry. Hence we gradually decrease $K$ by removing the $m$ furthest points each time until the ratio $\lambda_3^i$ by $\lambda_1^i$ (invariant of local sampling density and patch size) is smaller than a given threshold or a lower bound for $K$ is reached.

%\begin{algorithm}[h]
%\caption{Adaptive KNN}         
%\begin{algorithmic}
%\small 
%\STATE {Given maximum neighborhood size $K$, eigenvalue ratio threshold $r$ and shrink size $m$, for point $p_i \in P$ with initial neighborhood points set $\N_0(i)$ of size $K$}
%\STATE {\textbf{Initialization:} }
%\STATE {1.Compute local covariance matrix $\P_0$}
%\STATE {2.Comute $R = \frac{\lambda_{min}(P_0)}{\lambda_{max}(P_0)}$}
%\WHILE{$R \geq r$} 
%\STATE {1.Update $\N_{k+1}(i)$ by excluding furthest $m$ points from $\N_k(i)$} 
%\STATE {2.Update local covariance matrix $\P_{k+1}$}
%\STATE {3.Update $R_{k+1} = \frac{\lambda_{min}(P_{k+1})}{\lambda_{max}(P_{k+1})}$} 
%\ENDWHILE
%\end{algorithmic}
%\label{alg3} 
%\end{algorithm}

%\subsection{Adaptive-KNN based Local Mesh}

\noindent\textbf{Patch matching}~~
In real applications, well-sampled data for 3D shapes are not easy to obtain. Instead, holes, patches, or partial shapes are more common in real data. Correspondence between shapes with topological perturbations, artificial boundaries, and different sizes are difficult for methods based on global intrinsic descriptors in general.   
%of shape. Patch matching suffers from the fact that most patches have boundaries, and a pair of corresponding patches may have different sizes or some topological changes on the mesh. Traditional approaches dealing with shape correspondence, like 
%kernel matching \cite{vestner2017efficient} or spectrum generalized multi-dimensional scaling \cite{aflalo2016spectral}, since 
For example, the spectrum of LBO is sensitive to boundary conditions and topological changes.

However, since our method is based on local features, the effect of boundary conditions and topological perturbations are localized too. Hence our method can be directly applied to those scenarios with good performance. For example, our iterative method for the relaxed QAP using anchor pairs and sparsity control fits the smaller patch into the larger one nicely for partial matching (see Figure~\ref{patch3}). For post processing in patch matching, we switch from HKS to geodesic distance signature since HKS is sensitive to boundary conditions.% and topological changes.

\section{Experiment Results}
\label{sec:result}
We evaluate the performance of our method through various tests on data sets from TOSCA \cite{bronstein2008numerical} and SCAPE \cite{anguelov2005correlated} and on patches extracted from TOSCA. All inputs for our tests are raw data without any preprocessing, i.e., no low-resolution model or pre-computed geodesic distance. Experiments are conducted in Matlab on a PC with 16GB RAM and Intel i7-6800k CPU. The result of our method using mesh input is denoted as mesh method, and the result of our method without using mesh is denoted as point cloud method. 

\noindent\textbf{Error Metric}~~ Suppose our constructed correspondence maps $\x\in \M_1$ to $\y \in \M_2$ while the true correspondence is $\x$ is to $\y^*$, we measure the quality of our result by computing the geodesic error defined by $e(x) = \frac{d_{\M_2}(y,y^*)}{diam(\M_2)}$, where $diam(\M_2)$ is the geodesic diameter of $\M_2$. 
%True correspondence should produce geodesic error equal to 0. 

\begin{figure}
\centering\includegraphics[width=0.95\linewidth]{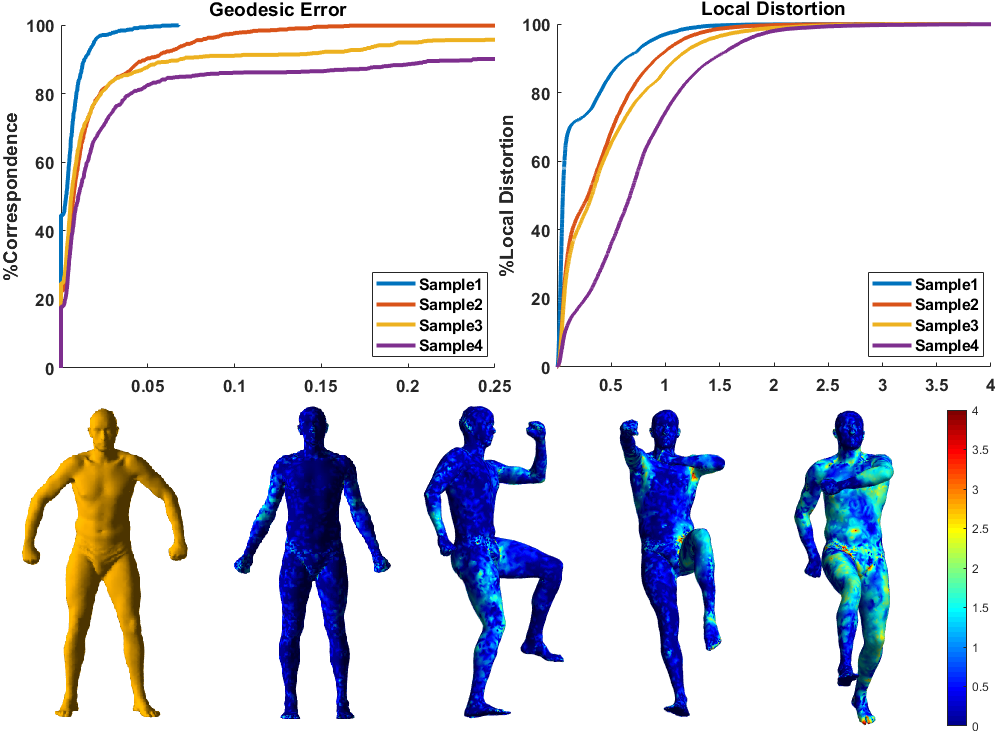}
\caption{Geodesic error and cumulative density function of local distortion for 4 sample pairs from the SCAPE data set (1st row); Original shape and corresponding local distortion heat map for sample 1 to 4 (2nd row). }
\label{faith}
\end{figure}

Local distortion defined in~\eqref{eq:faithfulness} can also serve as an unsupervised error metric to measure the quality of a map. As shown in Figure \ref{faith}, it's clear that local distortion decreases as the geodesic error decreases, which indicates that local distortion can serve as a good unsupervised metric to quantify the approximate isometry.

% \RJ{just my curiosity, or reviewers may also ask, how about our Faithfulness error for these data set? will add faithfulness error curve here}

\begin{figure}
%\begin{subfigure}[b]{0.5\linewidth}
\centering\includegraphics[width=.9\linewidth]{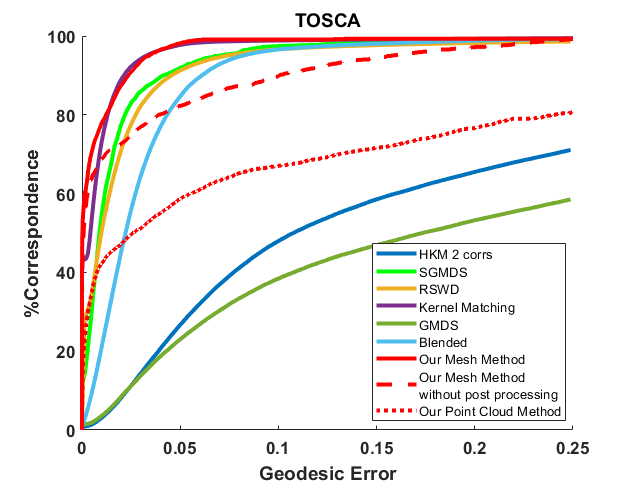}\\
%\end{subfigure}
%\begin{subfigure}[b]{0.5\linewidth}
\centering\includegraphics[width=.9\linewidth]{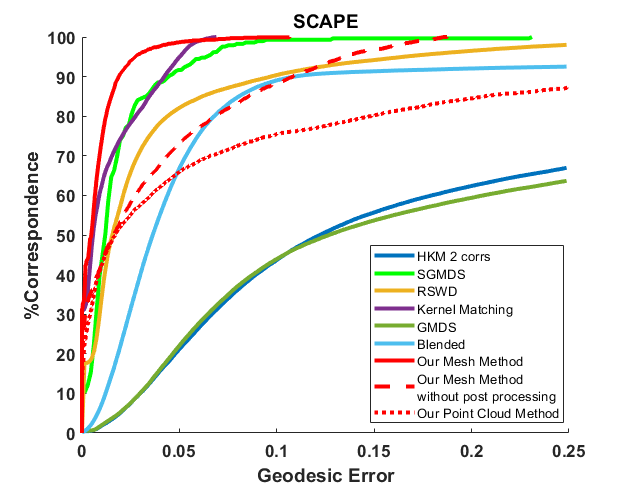}
%\end{subfigure}
\caption{Correspondence accuracy on the TOSCA and SCAPE data sets.}
\label{TSres}
\end{figure}
\noindent\textbf{TOSCA}~~ The TOSCA data set contains 76 shapes of 8 different classes, from humans to animals. The number of vertices varies from 4k to 50k.
We use 5 iterations to grow the set of anchor pairs. The neighborhood used for local distortion test for selecting anchor points is the second ring, and for sparsity control is the fourth ring. The distortion threshold decreases equally during the iterations from 5 to 1; the gradient descent step size in Algorithm~\ref{alg1} is 75; we approximate the heat kernel by 300 eigen-functions of the LBO with a diffusion time $t = 50$ in the post-processing step. For point clouds without mesh, we use an initial $K=200$, ratio $r=0.05$, and shrink size $m=6$ for our adaptive-KNN; HKS post processing is not used since the spectrum computed directly from the point cloud is not accurate enough. Results of our mesh method with or without post processing, and point cloud method without post processing are presented. We compare our method with the following methods: Blended \cite{kim2011blended}, SGMDS \cite{aflalo2016spectral}, GMDS \cite{bronstein2006generalized}, Kernel Marching \cite{vestner2017efficient}, RSWD \cite{lai2017multiscale}, and HKM 2 corrs \cite{ovsjanikov2010one}. Figure \ref{TSres} shows the quantitative result in terms of the geodesic error metric. Our mesh method outperforms most of the state-of-art methods. Our mesh method without post-processing and point cloud method also achieve reasonably good results. 

\noindent\textbf{SCAPE}~~ The SCAPE data set contains 72 shapes of humans in different poses. Each shape has 12,500 vertices. We use the same parameters as those on TOSCA data set except for diffusion time $t = 0.001$ in the post-processing step. Results of our mesh method with or without post processing, and point cloud method are presented. Figure \ref{TSres} shows the quantitative result. Our method achieves the state-of-art accuracy. Again, our mesh method without post-processing and point cloud method also achieve reasonably good results. 

\begin{figure}
\centering
\includegraphics[width=.9\linewidth]{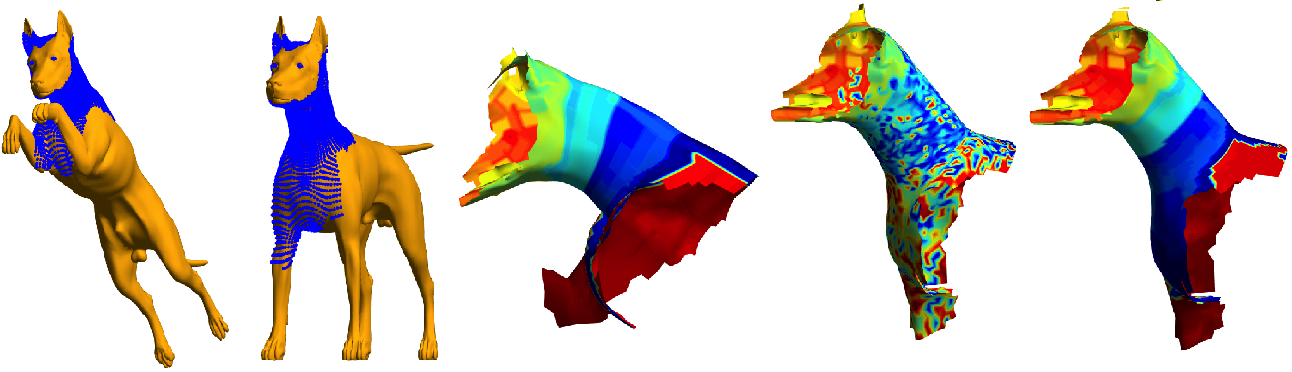}
\caption{Example of matching two patches. The first two images show the location of the patches; the third image is the color map on the first patch; the fourth image is the induced color map based on SHOT feature; the last image is the induced color map from our mesh method. 
% \RJ{rotate the dogs in the 1st row to match the view angles in the 2nd row. For the dogs in the 1st row, I suggest you use mesh for the patch part and make the rest part transparent. The same comments for figure 4.}
}
\label{patch_no_topo}
\end{figure}

\begin{figure}
\centering
\includegraphics[width=\linewidth]{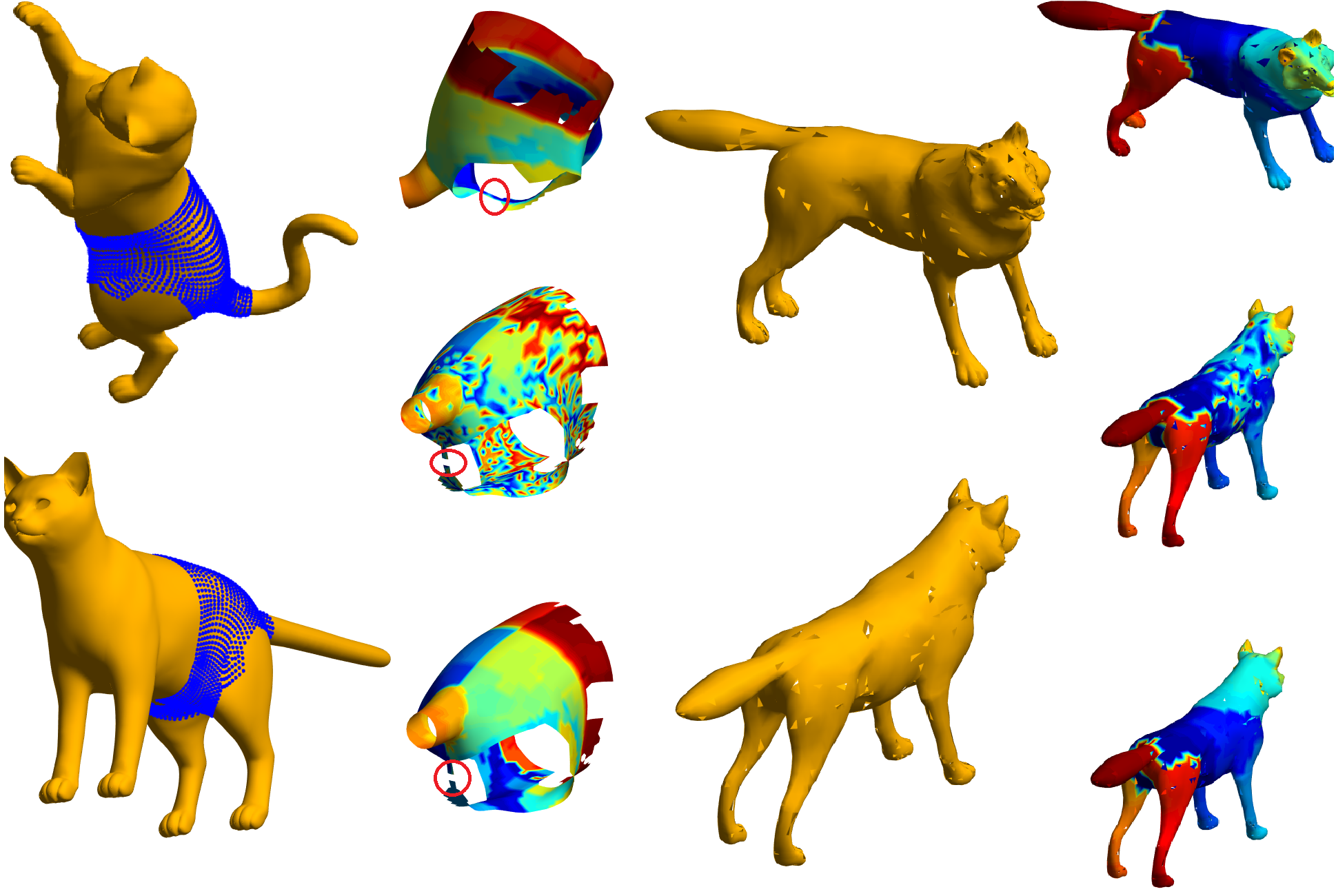}
\caption{Example of matching patches with topological perturbation and shapes with randomly missing elements. The first and third columns illustrate the patches and shapes to match. The top color map of the first patch/shape is mapped to the second patch/shape using SHOT (middle) and our method (bottom).}
\label{patch2}
\end{figure}

\noindent\textbf{Patch Matching}~~ We present a few test results for patches that have holes, boundaries, and partial matching. We paint the first patch with colors and map the color to the second patch with the correspondence computed using SHOT \cite{tombari2010unique} as the pointwise descriptor, which also serves as the initial guess for our method, and the correspondence computed from our algorithm to visualize the result. Since HKS is sensitive to boundary conditions and topological changes, we use geodesic distance to those selected anchor pairs as pointwise descriptor in the post processing step.

The first test is matching two patches of a dog with different poses from TOSCA, as shown in Figure \ref{patch_no_topo}. The two patches have very irregular boundaries. Using extrinsic pointwise descriptors, such as SHOT, fail to give a good dense correspondence. However, our method performs well since it uses local pairwise descriptors to find high-quality anchor pairs and integrates global pointwise descriptor, the geodesic distance to those anchor pairs, to complete the dense correspondence.  

The second test is matching two patches with topological perturbations from TOSCA data, as shown in Figure \ref{patch2}. The first case is two different poses of a wolf with mesh elements randomly deleted from each surface independently. The second case is body parts of a cat in different poses with topological perturbation, the second patch is not connected at the bottom while the first one is as highlighted in the figure.  Since neither local connectivity distortion nor missing elements will significantly influence the stiffness matrix or mass matrix at most points, our method can still produce good results.

We further test our method on a pair of patches with both different sizes (partial matching) and topological changes, as shown in Figure \ref{patch3}. The example is mapping an arm patch to an entire shape. Extra points are colored as blue. The arm patch has less points than entire shape and the figure tips are cut off, which results in both different size and topological change. Even for this challenging example, our method performs really well.

\noindent\textbf{Time efficiency}~~ We list the average run time of several examples in TOSCA data set in Table \ref{runtime}. Most state-of-the-art methods using (dense) pair-wise descriptors and quadratic assignment (QA) require dense matrix multiplication in each step which already has super-quadratic complexity. Although Laplace-Beltrami (LB) eigen-functions can be used to compress the dense matrix by low-rank approximation, it is still less sparse or localized and more time-consuming to compute than our simple, sparse and localized pair-wise descriptor. Combined with our sparsity-enforced method for QA, our method has at most $O(n^2)$  complexity which outperforms methods with super-quadratic complexity when handling data with large size. Experimentally, our method shows complexity even better than $O(n^2)$.

\begin{table}[]
\scalebox{0.8}{
\centering
\begin{tabular}{l|c|c|c|c|c}
\hline
Model                 & Wolf & Centaur & Horse & Cat   & David \\ \hline
Number of Vertices    & 4344 & 15768   & 19248 & 27894 & 52565 \\ \hline
Mesh Method(s)        & 59  & 531     & 801  & 929  & 1681  \\
Point Cloud Method(s) & 57  & 524     & 811  & 937  & 1610  \\ \hline
\end{tabular}}
\caption{Run time in seconds for examples from TOSCA data set.}
\label{runtime}
\end{table}

\section{Conclusion}
We develop a simple, effective iterative method to solve a relaxed quadratic assignment model through sparsity control for shape correspondence between two approximate isometric surfaces based on a novel local pairwise descriptor. Two key ideas of our iterative algorithm are: 1) select pairs with good correspondence as anchor points using a local unsupervised distortion test, 2) solve a regularized quadratic assignment problem only in the neighborhood of selected anchor points through sparsity control. With enough high-quality anchor points, various pointwise global features with reference to these anchor points can further improve the dense shape correspondence. Extensive experiments are conducted to show the efficiency, quality, and versatility of our method on large data sets, patches, and purely point cloud data.

Similar to many existing methods, our method will have
difficulty in dealing with significant non-isometric distortion
and highly non-uniform sampling. These will be further studied in our future research.

\section*{Acknowledgement}
R. Lai's research is supported in part by an NSF Career
Award DMS--1752934. H. Zhao is partially supported by NSF DMS-1418422. Part of this research was performed while the authors were visiting the Institute for Pure and Applied Mathematics (IPAM), which is supported by the NSF DMS-1440415.

{\small
\bibliographystyle{ieee_fullname}
\bibliography{egbib}

\begin{thebibliography}{10}\itemsep=-1pt

\bibitem{aflalo2014graph}
Yonathan Aflalo, Alex Bronstein, and Ron Kimmel.
\newblock Graph matching: relax or not?
\newblock {\em arXiv preprint arXiv:1401.7623}, 2014.

\bibitem{aflalo2015convex}
Yonathan Aflalo, Alexander Bronstein, and Ron Kimmel.
\newblock On convex relaxation of graph isomorphism.
\newblock {\em Proceedings of the National Academy of Sciences},
  112(10):2942--2947, 2015.

\bibitem{aflalo2016spectral}
Yonathan Aflalo, Anastasia Dubrovina, and Ron Kimmel.
\newblock Spectral generalized multi-dimensional scaling.
\newblock {\em International Journal of Computer Vision}, 118(3):380--392,
  2016.

\bibitem{anguelov2005correlated}
Dragomir Anguelov, Praveen Srinivasan, Hoi-Cheung Pang, Daphne Koller,
  Sebastian Thrun, and James Davis.
\newblock The correlated correspondence algorithm for unsupervised registration
  of nonrigid surfaces.
\newblock In {\em Advances in neural information processing systems}, pages
  33--40, 2005.

\bibitem{aubry2011wave}
Mathieu Aubry, Ulrich Schlickewei, and Daniel Cremers.
\newblock The wave kernel signature: A quantum mechanical approach to shape
  analysis.
\newblock In {\em 2011 IEEE international conference on computer vision
  workshops (ICCV workshops)}, pages 1626--1633. IEEE, 2011.

\bibitem{barzilai1988two}
Jonathan Barzilai and Jonathan~M Borwein.
\newblock Two-point step size gradient methods.
\newblock {\em IMA journal of numerical analysis}, 8(1):141--148, 1988.

\bibitem{berard1994embedding}
Pierre B{\'e}rard, G{\'e}rard Besson, and Sylvain Gallot.
\newblock Embedding riemannian manifolds by their heat kernel.
\newblock {\em Geometric \& Functional Analysis GAFA}, 4(4):373--398, 1994.

\bibitem{bronstein2006generalized}
Alexander~M Bronstein, Michael~M Bronstein, and Ron Kimmel.
\newblock Generalized multidimensional scaling: a framework for
  isometry-invariant partial surface matching.
\newblock {\em Proceedings of the National Academy of Sciences},
  103(5):1168--1172, 2006.

\bibitem{bronstein2008numerical}
Alexander~M Bronstein, Michael~M Bronstein, and Ron Kimmel.
\newblock {\em Numerical geometry of non-rigid shapes}.
\newblock Springer Science \& Business Media, 2008.

\bibitem{Bronstein:2010CVPR}
M.~M. Bronstein and I. Kokkinos.
\newblock Scale-invariant heat kernel signatures for non-rigid shape
  recognition.
\newblock {\em IEEE Conference on Computer Vision and Pattern Recognition
  (CVPR)}, pages 1704--1711, 2010.

\bibitem{bronstein2010scale}
Michael~M Bronstein and Iasonas Kokkinos.
\newblock Scale-invariant heat kernel signatures for non-rigid shape
  recognition.
\newblock In {\em 2010 IEEE Computer Society Conference on Computer Vision and
  Pattern Recognition}, pages 1704--1711. IEEE, 2010.

\bibitem{chavel1984eigenvalues}
Isaac Chavel.
\newblock {\em Eigenvalues in Riemannian geometry}, volume 115.
\newblock Academic press, 1984.

\bibitem{chen2015robust}
Qifeng Chen and Vladlen Koltun.
\newblock Robust nonrigid registration by convex optimization.
\newblock In {\em Proceedings of the IEEE International Conference on Computer
  Vision}, pages 2039--2047, 2015.

\bibitem{dubrovina2011approximately}
Anastasia Dubrovina and Ron Kimmel.
\newblock Approximately isometric shape correspondence by matching pointwise
  spectral features and global geodesic structures.
\newblock {\em Advances in Adaptive Data Analysis}, 3(01n02):203--228, 2011.

\bibitem{dym2017ds++}
Nadav Dym, Haggai Maron, and Yaron Lipman.
\newblock Ds++: A flexible, scalable and provably tight relaxation for matching
  problems.
\newblock {\em arXiv preprint arXiv:1705.06148}, 2017.

\bibitem{gasparetto2017spatial}
Andrea Gasparetto, Luca Cosmo, Emanuele Rodola, Michael Bronstein, and Andrea
  Torsello.
\newblock Spatial maps: From low rank spectral to sparse spatial functional
  representations.
\newblock In {\em 2017 International Conference on 3D Vision (3DV)}, pages
  477--485. IEEE, 2017.

\bibitem{gu2010discrete}
Xianfeng~David Gu, Ren Guo, Feng Luo, and Wei Zeng.
\newblock Discrete laplace-beltrami operator determines discrete riemannian
  metric.
\newblock {\em arXiv preprint arXiv:1010.4070}, 2010.

\bibitem{gumhold2001feature}
Stefan Gumhold, Xinlong Wang, and Rob~S MacLeod.
\newblock Feature extraction from point clouds.
\newblock In {\em IMR}. Citeseer, 2001.

\bibitem{halimi2018self}
Oshri Halimi, Or Litany, Emanuele Rodol{\`a}, Alex Bronstein, and Ron Kimmel.
\newblock Self-supervised learning of dense shape correspondence.
\newblock {\em arXiv preprint arXiv:1812.02415}, 2018.

\bibitem{kezurer2015tight}
Itay Kezurer, Shahar~Z Kovalsky, Ronen Basri, and Yaron Lipman.
\newblock Tight relaxation of quadratic matching.
\newblock In {\em Computer Graphics Forum}, volume~34, pages 115--128. Wiley
  Online Library, 2015.

\bibitem{kim2011blended}
Vladimir~G Kim, Yaron Lipman, and Thomas Funkhouser.
\newblock Blended intrinsic maps.
\newblock In {\em ACM Transactions on Graphics (TOG)}, volume~30, page~79. ACM,
  2011.

\bibitem{lai2013local}
Rongjie Lai, Jiang Liang, and Hongkai Zhao.
\newblock A local mesh method for solving pdes on point clouds.
\newblock {\em Inverse Problems \& Imaging}, 7(3), 2013.

\bibitem{lai2010metric}
Rongjie Lai, Yonggang Shi, Kevin Scheibel, Scott Fears, Roger Woods, Arthur~W
  Toga, and Tony~F Chan.
\newblock Metric-induced optimal embedding for intrinsic 3d shape analysis.
\newblock In {\em 2010 IEEE Computer Society Conference on Computer Vision and
  Pattern Recognition}, pages 2871--2878. IEEE, 2010.

\bibitem{lai2017multiscale}
Rongjie Lai and Hongkai Zhao.
\newblock Multiscale nonrigid point cloud registration using rotation-invariant
  sliced-wasserstein distance via laplace--beltrami eigenmap.
\newblock {\em SIAM Journal on Imaging Sciences}, 10(2):449--483, 2017.

\bibitem{lawler1963quadratic}
Eugene~L Lawler.
\newblock The quadratic assignment problem.
\newblock {\em Management science}, 9(4):586--599, 1963.

\bibitem{leordeanu2005spectral}
Marius Leordeanu and Martial Hebert.
\newblock A spectral technique for correspondence problems using pairwise
  constraints.
\newblock In {\em Tenth IEEE International Conference on Computer Vision
  (ICCV'05) Volume 1}, volume~2, pages 1482--1489. IEEE, 2005.

\bibitem{Levy:2006IEEECSMA}
B. Levy.
\newblock Laplace-beltrami eigenfunctions: Towards an algorithm that
  understands geometry.
\newblock {\em IEEE International Conference on Shape Modeling and
  Applications, invited talk}, 2006.

\bibitem{memoli2011gromov}
Facundo M{\'e}moli.
\newblock Gromov--wasserstein distances and the metric approach to object
  matching.
\newblock {\em Foundations of computational mathematics}, 11(4):417--487, 2011.

\bibitem{ovsjanikov2012functional}
Maks Ovsjanikov, Mirela Ben-Chen, Justin Solomon, Adrian Butscher, and Leonidas
  Guibas.
\newblock Functional maps: a flexible representation of maps between shapes.
\newblock {\em ACM Transactions on Graphics (TOG)}, 31(4):30, 2012.

\bibitem{ovsjanikov2010one}
Maks Ovsjanikov, Quentin M{\'e}rigot, Facundo M{\'e}moli, and Leonidas Guibas.
\newblock One point isometric matching with the heat kernel.
\newblock In {\em Computer Graphics Forum}, volume~29, pages 1555--1564. Wiley
  Online Library, 2010.

\bibitem{reddy1993introduction}
Junuthula~Narasimha Reddy.
\newblock An introduction to the finite element method.
\newblock {\em New York}, 1993.

\bibitem{reuter2009discrete}
Martin Reuter, Silvia Biasotti, Daniela Giorgi, Giuseppe Patan{\`e}, and
  Michela Spagnuolo.
\newblock Discrete laplace--beltrami operators for shape analysis and
  segmentation.
\newblock {\em Computers \& Graphics}, 33(3):381--390, 2009.

\bibitem{Reuter:06}
M. Reuter, F.E. Wolter, and N. Peinecke.
\newblock Laplace-{B}eltrami spectra as {S}hape-{DNA} of surfaces and solids.
\newblock {\em Computer-Aided Design}, 38:342--366, 2006.

\bibitem{rodola2012game}
Emanuele Rodola, Alex~M Bronstein, Andrea Albarelli, Filippo Bergamasco, and
  Andrea Torsello.
\newblock A game-theoretic approach to deformable shape matching.
\newblock In {\em 2012 IEEE Conference on Computer Vision and Pattern
  Recognition}, pages 182--189. IEEE, 2012.

\bibitem{sahni1976p}
Sartaj Sahni and Teofilo Gonzalez.
\newblock P-complete approximation problems.
\newblock {\em Journal of the ACM (JACM)}, 23(3):555--565, 1976.

\bibitem{schonsheck2018nonisometric}
Stefan~C Schonsheck, Michael~M Bronstein, and Rongjie Lai.
\newblock Nonisometric surface registration via conformal laplace-beltrami
  basis pursuit.
\newblock {\em arXiv preprint arXiv:1809.07399}, 2018.

\bibitem{sun2009concise}
Jian Sun, Maks Ovsjanikov, and Leonidas Guibas.
\newblock A concise and provably informative multi-scale signature based on
  heat diffusion.
\newblock In {\em Computer graphics forum}, volume~28, pages 1383--1392. Wiley
  Online Library, 2009.

\bibitem{tevs2011intrinsic}
Art Tevs, Alexander Berner, Michael Wand, Ivo Ihrke, and H-P Seidel.
\newblock Intrinsic shape matching by planned landmark sampling.
\newblock In {\em Computer Graphics Forum}, volume~30, pages 543--552. Wiley
  Online Library, 2011.

\bibitem{tombari2010unique}
Federico Tombari, Samuele Salti, and Luigi Di~Stefano.
\newblock Unique signatures of histograms for local surface description.
\newblock In {\em European conference on computer vision}, pages 356--369.
  Springer, 2010.

\bibitem{Vallet:2008CGF}
B. Vallet and B. Levy.
\newblock Spectral geometry processing with manifold harmonics.
\newblock {\em Computer Graphics Forum (Proceedings Eurographics)}, 2008.

\bibitem{van2011survey}
Oliver Van~Kaick, Hao Zhang, Ghassan Hamarneh, and Daniel Cohen-Or.
\newblock A survey on shape correspondence.
\newblock In {\em Computer Graphics Forum}, volume~30, pages 1681--1707. Wiley
  Online Library, 2011.

\bibitem{vestner2017efficient}
Matthias Vestner, Zorah L{\"a}hner, Amit Boyarski, Or Litany, Ron Slossberg,
  Tal Remez, Emanuele Rodola, Alex Bronstein, Michael Bronstein, Ron Kimmel,
  et~al.
\newblock Efficient deformable shape correspondence via kernel matching.
\newblock In {\em 3D Vision (3DV), 2017 International Conference on}, pages
  517--526. IEEE, 2017.

\bibitem{vestner2017product}
Matthias Vestner, Roee Litman, Emanuele Rodol{\`a}, Alex Bronstein, and Daniel
  Cremers.
\newblock Product manifold filter: Non-rigid shape correspondence via kernel
  density estimation in the product space.
\newblock In {\em Proceedings of the IEEE Conference on Computer Vision and
  Pattern Recognition}, pages 3327--3336, 2017.

\bibitem{wang2011discrete}
Chaohui Wang, Michael~M Bronstein, Alexander~M Bronstein, and Nikos Paragios.
\newblock Discrete minimum distortion correspondence problems for non-rigid
  shape matching.
\newblock In {\em International Conference on Scale Space and Variational
  Methods in Computer Vision}, pages 580--591. Springer, 2011.

\end{thebibliography}
}

\end{document}